\documentclass{article}

\usepackage[preprint]{neurips_2025}

\usepackage[utf8]{inputenc} 
\usepackage[T1]{fontenc}    
\usepackage{hyperref}       
\usepackage{url}            
\usepackage{booktabs}       
\usepackage{amsfonts}       
\usepackage{nicefrac}       
\usepackage{microtype}      
\usepackage{xcolor}         
\usepackage{graphicx}
\usepackage{multirow}
\usepackage{algorithm}
\usepackage{algpseudocode}
\usepackage{wrapfig}
\usepackage{enumitem}

\usepackage[acronym,shortcuts]{glossaries}
\newacronym{BIRD}{BIRD}{Behavior Induction via Representation-structure Distillation}
\newacronym{CKA}{CKA}{Centered Kernel Alignment}
\newacronym{AI}{AI}{artifical intelligence}
\newacronym{OOD}{OOD}{out-of-distribution}
\newacronym{KD}{KD}{knowledge distillation}
\newacronym{LwF}{LwF}{Learning without Forgetting}
\newacronym{RSA}{RSA}{Representational Similarity Analysis}
\newacronym{PGR}{PGR}{Performance Gap Recovered}

\glsdisablehyper  

\title{BIRD: Behavior Induction via Representation-structure Distillation}

\author{%
    Galen Pogoncheff\\
    Department of Computer Science\\
    UC Santa Barbara\\
    \texttt{galenpogoncheff@ucsb.edu} \\
    \And
    Michael Beyeler\\
    Department of Computer Science\\
    Department of Psychological \& Brain Sciences\\
    UC Santa Barbara\\
    \texttt{mbeyeler@ucsb.edu} \\
}

\begin{document}

\maketitle

\begin{abstract}

Human-aligned deep learning models exhibit behaviors consistent with human values, such as robustness, fairness, and honesty.
Transferring these behavioral properties to models trained on different tasks or data distributions remains challenging: aligned behavior is easily forgotten during fine-tuning, and collecting task-specific data that preserves this behavior can be prohibitively costly.
We introduce BIRD (Behavior Induction via Representation-structure Distillation), a flexible framework for transferring aligned behavior by matching the internal representation structure of a student model to that of a teacher.
Applied to out-of-distribution robustness in image classification, BIRD outperforms fine-tuning, transfer learning, and continual learning methods, improving robust accuracy by up to 16\% over the next strongest baseline. It remains effective even when the teacher is trained on a much simpler dataset and is $25 \times$ smaller than the student.
In a large-scale study of over 400 teacher-student pairs, we show that three interpretable and computable properties of the teacher's representations (i.e., task relevance, behavioral relevance, and complementary knowledge) explain up to 85\% of the variance in transfer success.
These insights offer practical guidance for teacher selection and design.
BIRD turns small, well-aligned models into scalable alignment seeds, removing a key bottleneck in deploying safe AI systems in the wild.

\end{abstract}

\section{Introduction}

As AI systems become increasingly capable, their alignment with human values (expressed through traits like robustness, fairness, and controllability) has become a central challenge~\citep{safe_ai_deepmind_2018, ji2023ai}.
Behavioral alignment typically requires costly supervision: adversarial training, human feedback, or special-purpose datasets \citep{madry2017towards,ouyang2022training, rafailov2023direct}.
These techniques do not easily scale to new tasks or domains.

A natural goal is to transfer aligned behavior from one model to another.
Yet behavior often degrades during fine-tuning \citep{shafahi2019adversarially,qi2023fine,peng2024navigating}, and most transfer methods assume the teacher and student share training data (sometimes unlabeled) or a common output space \citep{burns2023weak,zhou2025weak}.
Worse, the datasets used to train aligned models (especially those involve human feedback) are often private or impractical to share.

In the context of adversarial robustness, this issue has been addressed by fine-tuning robust feature encoders \citep{shafahi2019adversarially, nern2023transfer} or by incorporating learning strategies that mitigate catastrophic forgetting \citep{shafahi2019adversarially, fan2021does}.
However, these approaches typically assume that robust features will generalize across domains, which only holds with sufficiently large and diverse pre-training datasets, which are costly to obtain and train on.

Recent work in weak-to-strong generalization offers a promising alternative: train a small, well-aligned ``weak'' model and use it to supervise a larger ``strong'' model \citep{burns2023weak}.
However, these approaches still assume the teacher and student share a task, training dataset, or output domain. 
In this work, we ask: \emph{Can aligned behavior be transferred even when the teacher and student differ in architecture, task, and training data?}
We investigate this question by analyzing which properties of a teacher's representation space enable aligned behavior transfer across heterogeneous models.

Here, we introduce BIRD (Behavior Induction via Representation-structure Distillation), a simple, drop-in framework for transferring aligned behavior between heterogeneous models by distilling task- and behaviorally-relevant structure from the teacher's representation space into the student's.
BIRD requires no access to teacher training data and succeeds even when the teacher is trained on simpler tasks or domains, enabling scalable reuse of aligned models.

BIRD is inspired by recent work in NeuroAI, where researchers hypothesize that desirable behavioral properties (e.g., robustness to noise, invariance across transformations) may be encoded in the geometry of brain representations \citep{chung2021neural, zador2023catalyzing,mineault2024neuroai}.
Studies biasing deep neural networks to learn representations similar to those observed in neural data have shown improved generalization and robustness on image classification benchmarks \citep{dapello2023aligning, safarani2021towards, li2019brainlearning}. 
However, these gains are limited to the data distribution of the neural data training set, and degrade when transferring to categories or domains outside of this set \citep{dapello2023aligning}. 
Moreover, such work typically requires expensive brain recordings and lacks actionable criteria for selecting alignment layers or teacher representations.

Our work builds on the core insight that behavioral properties are encoded in the structure of a model’s latent representations \citep{zou2023representation}, but generalizes it in two key ways. First, BIRD does not rely on biological recordings or shared datasets; the teacher and student may differ in size, architecture, domain, and output space. Second, we empirically identify three computable representation properties (i.e., task relevance, behavioral alignment, and complementary knowledge) that reliably predict transfer success. This enables principled selection of both teacher models and representation layers.

We evaluate BIRD in the context of transferring out-of-distribution robustness in image classification, spanning over 400 teacher-student pairs that vary in architecture, capacity, and training data. We find:
\begin{itemize}[topsep=0pt,leftmargin=15pt,parsep=0pt]
    \item BIRD outperforms existing transfer methods including fine-tuning, continual learning, and activation-based distillation, improving robust accuracy by up to 16\% over the strongest baseline.
    \item BIRD enables weak-to-strong transfer from small, simple teachers (e.g., CIFAR-10-trained MobileNetV2) to students up to $25\times$ larger, trained on more complex datasets such as TinyImageNet.
    \item Transfer success is predictable and actionable: three interpretable properties of the teacher's representation space (i.e., task relevance, behavioral relevance, and complementary knowledge) explain up to 85\% of the variance in transfer outcomes, offering practical guidance for layer and teacher selection.
\end{itemize}

By aligning representation structures instead of activation values or model outputs, BIRD provides a flexible, data-free mechanism for transferring aligned behavior across models and domains. 
It advances the promise of weak-to-strong generalization and sets the stage for scalable, reusable alignment across tasks.

\section{Related work}

\subsection{Scalable alignment via Weak-to-Strong supervision}

Behavioral misalignment arises when models optimize their training objective while diverging from human intent \citep{amodei2016concrete, di2022goal, razin2024unintentional}. 
To mitigate this, researchers often rely on special-purpose datasets and training strategies such as preference tuning for language models \citep{ouyang2022training, rafailov2023direct} or adversarial training for vision models \citep{madry2017towards, hendrycks2021many}. 
While effective, these approaches are difficult to scale due to the need for curated data and costly supervision.

\emph{Scalable oversight} aims to reduce this dependence by developing methods for supervising advanced models efficiently, even when human feedback is limited or unavailable \citep[see][]{ji2023ai}. 
A promising direction is weak-to-strong generalization, in which a small, well-aligned model provides soft-label supervision to guide the training of a larger, more capable model \citep{burns2023weak, zhu2024weak, zhou2025weak}.
This reduces the need for direct human supervision, but typically assumes that teacher and student operate on the same task, output space, and dataset.

Our work generalizes this paradigm. 
Instead of requiring shared output spaces and access to a teacher's training data (even if not labeled), we transfer aligned behavior between heterogeneous models by leveraging the representation structure of a teacher model as a supervisory signal. 
This enables weak-to-strong transfer across differing tasks, datasets, and architectures.

\subsection{Limits of robustness transfer in image models}

Adversarial robustness has been a major benchmark for studying behavioral transfer \citep{shafahi2019adversarially, chen2021cartl, nern2023transfer, liu2023twins, xu2023enhancing}. 
Classical approaches require on-the-fly generation of adversarial examples during training \citep{madry2017towards, schmidt2018adversarially}, making them computationally expensive. 
Recent work has explored transferring robustness to new tasks without retraining from scratch.

A key challenge is \emph{catastrophic forgetting} \citep{kirkpatrick2017overcoming}.
When robust models are fine-tuned on new data, robustness is often lost \citep{shafahi2019adversarially, nern2023transfer}. 
To address this, methods have incorporated adversarial examples during transfer or added constraints to limit feature drift \citep{chen2021cartl, liu2023twins, xu2023enhancing}.

Shafahi et al. \citep{shafahi2019adversarially} proposed adversarially robust transfer learning using \ac{LwF} \citep{li2017learning}, which preserves robustness by constraining changes to the final-layer features. 
However, these techniques assume that robust features generalize across domains, which holds only with large and diverse pretraining datasets. 
In contrast, our work enables behavior transfer from smaller, simpler models trained on low-resource domains, without requiring shared inputs or labels.

\subsection{Distilling representation geometry}

\emph{\Ac{KD}} enables a student model to learn from a teacher by mimicking outputs or hidden activations. 
Originally developed for model compression \citep{Bucilua, hinton2015distilling}, KD has since expanded to include intermediate-layer supervision \citep{romero2014fitnets, zagoruyko2016paying} and cross-modal transfer \citep{gupta2016cross}.

Standard \ac{KD} assumes shared tasks and output spaces. More recent approaches distill from internal activations, guiding the student to match the teacher’s intermediate features \citep{romero2014fitnets, zagoruyko2016paying}.
These approaches transfer information at the level of individual examples or layers.

BIRD takes a different approach. 
Rather than matching outputs or activations, we transfer the pairwise structure of the teacher's representation space, captured via Gram matrices over batches of inputs. 
This quantifies the geometry of internal representations, rather than specific activation values.

This builds on the idea that a model's knowledge lies not only in its outputs, but in the organization of its representation space \citep{hjelm2018learning, tian2019contrastive, muttenthaler2024aligning}. 
While prior work uses this idea for unsupervised learning or intra-task alignment, BIRD applies it to cross-task, cross-domain behavior transfer, generalizing \ac{KD} into a scalable mechanism for aligned supervision.

\subsection{Representation alignment in NeuroAI}

Work in NeuroAI suggests that robust and general behavior in biological systems may arise from the structure of neural representations \citep{chung2021neural, zador2023catalyzing}. 
Several studies have attempted to bias AI models toward brain-like representations by minimizing dissimilarity to neural recordings from visual cortex, often using objectives based on \ac{CKA} or \ac{RSA} \citep{dapello2023aligning, safarani2021towards, muttenthaler2024aligning}. 
These approaches have shown moderate improvements in robustness but typically require neural data, assume shared input domains, and show limited generalization beyond the training set.

Our work draws inspiration from these ideas but removes the need for brain recordings or stimulus overlap. 
BIRD operationalizes the hypothesis that structured representations support general behavior, using any aligned model as a teacher. 
By enabling behavior transfer across architectures and domains, BIRD extends NeuroAI insights into a general-purpose framework for scalable alignment.

\section{BIRD: Behavior Induction via Representation-structure Distillation}
\label{sec:bird_methods}

\begin{figure}[t!]
  \centering
  \includegraphics[width=\linewidth]{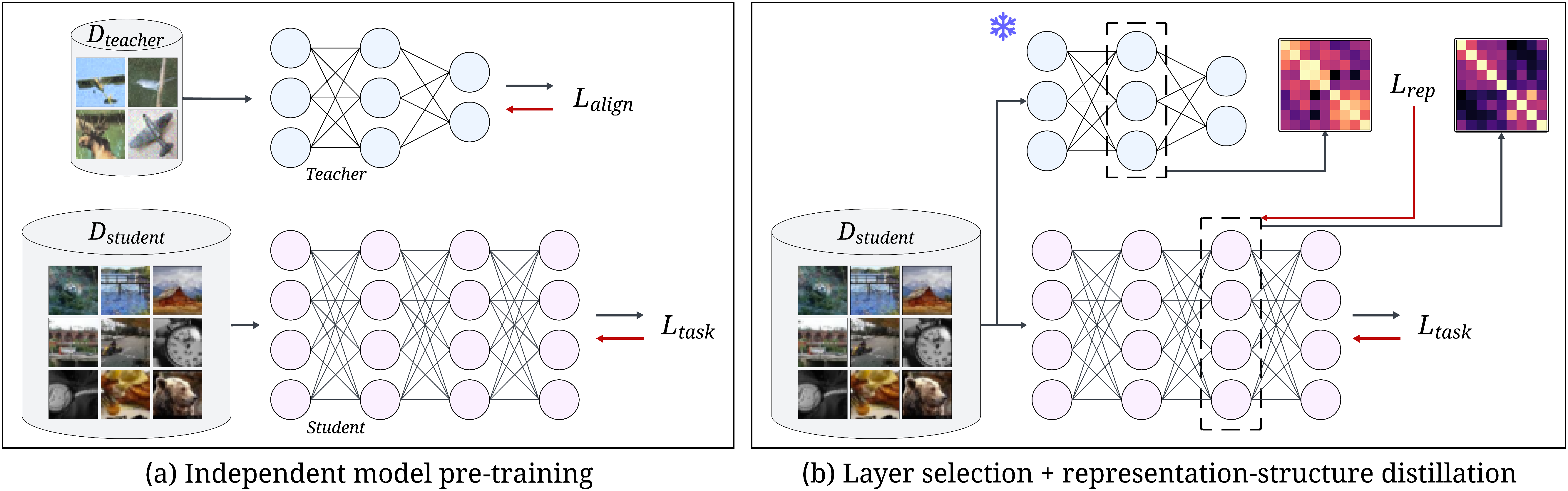}
  \caption{Overview of BIRD: \textbf{(a)} First, a student model is pre-trained using generic optimization approaches on its target training set, $\mathcal{D}_\mathrm{student}$. 
  A teacher model is independently trained to develop aligned behavior (e.g., robustness) on dataset $\mathcal{D}_\mathrm{teacher}$ (optimizing an alignment objective, $\mathcal{L}_{\text{align}}$). \textbf{(b)} Next, the student model is fine-tuned on its original dataset to maintain task performance while learning latent representations with similar structure to that of the (frozen) teacher in selected layers.}
  \label{fig:bird_overview}
\end{figure}

We introduce \ac{BIRD}, a flexible framework for transferring aligned behavior from a teacher model to a student model.
Our approach is grounded in the hypothesis that task-general behavioral properties (i.e., robustness, fairness, and invariance) are encoded in the structure of a model's internal representation space, as suggested by recent work \citep{zou2023representation}. 
We posit that guiding a student model to adopt similar representational structure to its teacher biases it toward learning the same aligned behaviors.

BIRD proceeds in three steps (Figure~\ref{fig:bird_overview}):
\begin{enumerate}[topsep=0pt,leftmargin=15pt,parsep=0pt]
    \item We assume access to a trained teacher model $g_{\phi}: \mathcal{D}_\mathrm{teacher} \rightarrow \mathcal{Y}_\mathrm{teacher}$ that exhibits desirable behavioral properties, and a pretrained student model $f_{\theta}: \mathcal{D}_\mathrm{student} \rightarrow \mathcal{Y}_\mathrm{student}$ for which we wish to induce those properties (Figure~\ref{fig:bird_overview}a).
    \item A guiding layer in the teacher and a guided layer in the student are selected for distillation based on their relevance to task and behavioral alignment (Section~\ref{sec:explaining_transfer}).
    \item The student is fine-tuned to preserve performance on its original task while learning a representation space whose structure mimics that of the teacher (Figure~\ref{fig:bird_overview}b).
\end{enumerate}

To implement this, we define a loss that combines task performance and representational alignment:
\begin{equation}
    \label{eq:bird_loss}
    \mathbb{E}_{B \sim \mathcal{D}_\mathrm{student}} \Big[ \alpha  \mathcal{L}_\mathrm{task} \big( f_{\theta}(B), \cdot \big) + (1-\alpha) \mathcal{L}_\mathrm{rep} \big( u(B), v(B) \big) \Big]
\end{equation}
Here, $B$ is a batch of inputs from the student's training distribution and $\alpha$ is a hyperparameter that weighs the relative contributions of task and representation-structure loss.
The functions $u$ and $v$ map those inputs to intermediate layer representations in the teacher and student, respectively. 
The first term, $\mathcal{L}_\mathrm{task}$, is the task-specific loss that the student was originally trained to minimize (e.g., cross-entropy).
The second term, $\mathcal{L}_\mathrm{rep}$, penalizes dissimilarity in representation structure:
\begin{equation}
    \mathcal{L}_{rep}(u(B), v(B)) = 1-\text{CKA}_{\text{linear}}(u(B), v(B)).
\end{equation}

We use \ac{CKA}~\citep{kornblith2019similarity} to quantify the alignment of pairwise similarity structures within a batch of inputs:
\begin{equation}
    \label{cka_linear}
    \text{CKA}_{\text{linear}}(u(B), v(B)) = \frac{\vert \vert v(B)^T u(B)\vert \vert ^2_\text{F}}{\vert \vert u(B)^T u(B)\vert \vert ^2_\text{F} \cdot \vert \vert v(B)^T v(B)\vert \vert ^2_\text{F}}.
\end{equation}
We select \ac{CKA} because (i) it has proven effective at comparing internal representations in deep networks, (ii) it provides reliable similarity estimates for high-dimensional representations, and (iii) it is easy to interpret.

Our approach draws inspiration from recent work in neuroscience-informed representation learning, where deep networks are trained to jointly minimize a task loss and a neural alignment loss based on brain recordings \citep{li2019brainlearning, federer2020improved, safarani2021towards, dapello2023aligning}. 
However, unlike those approaches, BIRD does not require neural data or shared stimulus distributions.

Crucially, BIRD does not require access to the teacher's training data. 
The only inputs seen by the teacher are drawn from the student's domain, and the teacher provides supervision by projecting those inputs into its own representation space. 
Thus, behavioral alignment is induced without direct exposure to a behavioral ground truth signal (e.g., adversarial examples or human labels).

A key novelty of BIRD is that the teacher and student need not share a dataset, label space, or task objective. 
This sets our work apart from prior approaches to weak-to-strong generalization, which assume shared objectives and output spaces between teacher and student as well as use of the teachers training data \citep{burns2023weak,zhou2025weak}. 
By supervising over representational structure instead of activation values or predictions, BIRD supports transfer across heterogeneous models and domains.

\begin{table}[t]
\caption{Accuracy (\%) of MobileNetV2 (MN2), ResNet18 (RN18), and DenseNet169 (DN169) models after behavior transfer using clean data from  CIFAR-10 (C10), CIFAR-100 (C100), and TinyImageNet (TIN).  Values reported are accuracy over all clean and corrupted images (for held-out corruption types ``speckle-noise'', ``gaussian-blur'', ``spatter'', and ``saturate'' and corruption severities $1$-$5$) from the target test set, averaged over 3 seeds.} \label{tab:robustness}
\centering
\begin{tabular}{lcc c c c c c c c}
\toprule
 Model & Source & Target & \multicolumn{7}{c}{Accuracy of Behavior Transfer Method ($\uparrow$)} \\
 \cmidrule(lr){4-10}
 & Data & Data & None & LP & FT & LP-FT & Hints & LwF &  BIRD \\
\midrule
\multirow{3}{*}{MN2} & C10  & C100 & 51.31 & 10.95 & 51.12 & 47.56 & 51.53 & 52.21 & \textbf{54.77}  \\
                             & C10  & TIN  & 20.74 & 5.24  & 20.00 & 18.12 & 21.27 & 20.52 & \textbf{24.11}  \\
                             & C100 & TIN  & 20.74 & 18.84 & 20.66 & 23.52 & 21.26 & 23.18 & \textbf{25.03}  \\
\midrule
\multirow{3}{*}{RN18} & C10 & C100  & 52.03 & 16.93 & 51.95  & 50.63 & 52.20 & 55.42 & \textbf{57.39}  \\
                          & C10 & TIN  & 20.56 & 7.25  & 20.10 & 19.43 & 20.92 & 22.17 & \textbf{23.60}  \\
                          & C100 & TIN & 20.56 & 20.95 & 20.75 & 23.66 & 20.71 & 24.48 & \textbf{24.49}  \\
\midrule
\multirow{3}{*}{DN169} & C10 & C100 & 54.51 & 23.92 & 55.84 & 53.39 & 54.92 & 56.92 & \textbf{59.04}  \\
                             & C10 & TIN  & 22.59 & 10.66 & 23.39 & 21.20 & 22.68 & 24.14 & \textbf{25.25}  \\
                             & C100 & TIN & 22.59 & 23.55 & 23.19 & 24.86 & 22.75 & 26.14 & \textbf{27.46}  \\
\bottomrule \\
\end{tabular}
\end{table}

\section{Transferring robustness across datasets and architectures}
\label{sec:bird_robust_transfer}

\begin{figure}[b!]
  \centering
  \includegraphics[width=0.92\linewidth]{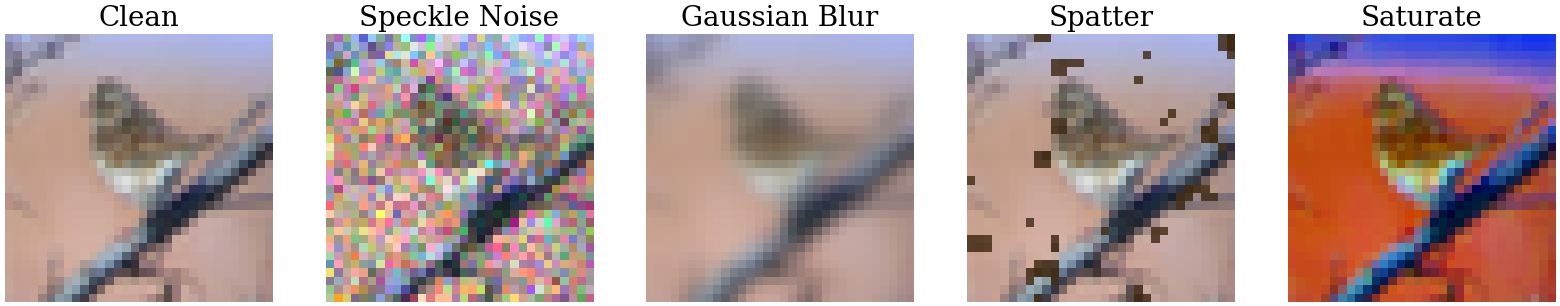}
  \caption{Sample image from CIFAR-10 with test-set corruptions from the ImageNet-C benchmark.  Corruptions are shown at highest severity (5) for clarity.}
  \label{fig:corrupt}
\end{figure}

We first evaluate BIRD in the context of robust image classification, where the goal is to transfer \ac{OOD} robustness from a teacher model trained on a lower-complexity dataset $\mathcal{D}_\mathrm{teacher}$ to a student model trained on clean data from a higher-complexity dataset $\mathcal{D}_\mathrm{student}$, where $\mathcal{D}_\mathrm{student} \ne \mathcal{D}_\mathrm{teacher}$.
This setting aligns well with the aims of weak-to-strong generalization, where aligned behavior is induced in a larger, less specialized model using supervision from a smaller, aligned one.

\paragraph{Setup and evaluation protocol}
We consider three dataset pairs: CIFAR-10 $\rightarrow$ CIFAR-100, CIFAR-10 $\rightarrow$ TinyImageNet, and CIFAR-100 $\rightarrow$ TinyImageNet \citep{cifarkrizhevsky2009learning,deng2009imagenet}. Teachers are trained to be robust to 15 ImageNet-C corruptions \citep{hendrycks2019benchmarking}, while students see only clean images.

Robustness is measured using accuracy over the clean and corrupted test set and \ac{PGR}~\citep{burns2023weak}:
\begin{equation}
    \label{eq:pgr}
    \text{PGR} = \min\left( \max\left(0, \frac{\text{Acc}_{\text{post}} - \text{Acc}_{\text{pre}}}{\text{Acc}_{\text{ceiling}} - \text{Acc}_{\text{pre}}} \right), 1 \right),
\end{equation}
where $\text{Acc}_{\text{post}}$ is student accuracy after behavior transfer, $\text{Acc}_{\text{pre}}$ is the pre-transfer baseline, and $\text{Acc}_{\text{ceiling}}$ is the accuracy of a student trained directly with access to OOD corruptions.

We test BIRD across three architectures: MobileNetV2 \citep{sandler2018mobilenetv2}, ResNet18 \citep{he2016deep}, and DenseNet169 \citep{huang2017densely}.
We compare BIRD to five baseline strategies that do not access corrupted target training data: linear probing (LP), full fine-tuning (FT), sequential LP followed by FT (LP-FT) \citep{kumar2022fine,nern2023transfer}, \ac{LwF} \citep{li2017learning,shafahi2019adversarially}, and hint-based distillation (Hints) \citep{romero2014fitnets}, which aligns activation values between teacher and student via linear mapping and mean-squared error representation loss.
Additional training details, corruption visualizations (a subset of which are shown in Figure \ref{fig:corrupt}), and evaluation breakdowns are provided in Appendix~\ref{app:robust_transfer}.

\paragraph{Comparison with baselines}
Across all dataset pairs and model architectures, BIRD achieves the highest out-of-distribution robustness and PGR (Table~\ref{tab:robustness}). 
For instance, when transferring robustness from CIFAR-10 to CIFAR-100, BIRD, on average, improves robustness by 4.5 percentage points and recovers 32.6\% of the performance gap to the robustness ceiling. The next best method, \ac{LwF}, recovers only 16.2\%. Similar trends are observed in the CIFAR-100 $\rightarrow$ TinyImageNet (23.4\% vs. 17.8\%) and CIFAR-10 $\rightarrow$ TinyImageNet (16.2\% vs. 5.8\%) transfers. LP and FT alone fail to consistently improve robustness over the baseline, suggesting that robust features learned on simpler datasets often do not generalize to more complex ones and are easily forgotten when fine-tuned.

LP-FT provides modest gains, but only in the CIFAR-100 $\rightarrow$ TinyImageNet transfer. This highlights a key challenge of behavior transfer from weak sources: robust but highly specific features may not generalize across distribution shifts unless structural constraints are imposed during transfer.

\begin{figure}[t!]
  \centering
  \includegraphics[width=0.95\linewidth]{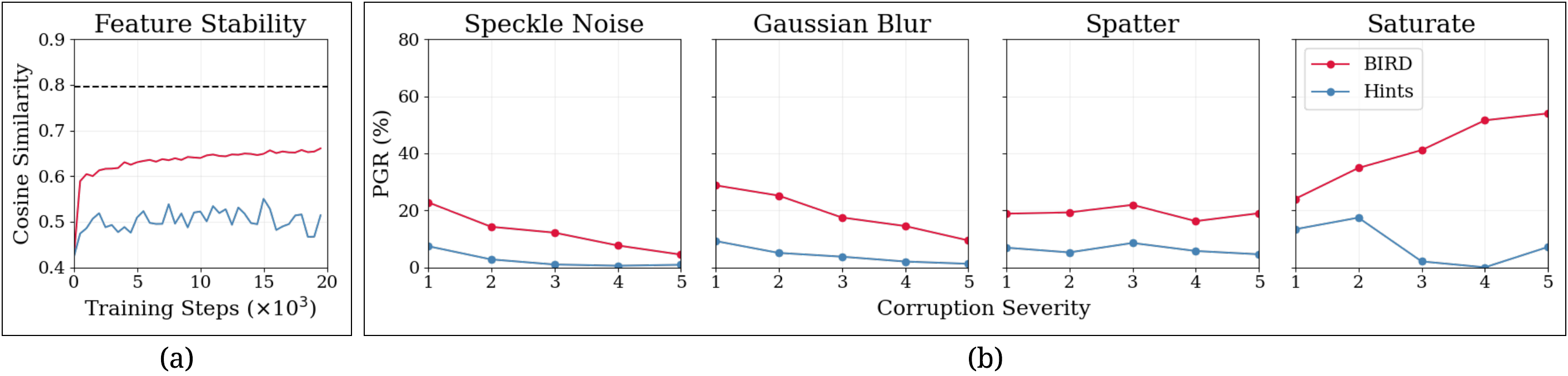}
  \caption{Comparing transferred robustness between BIRD and Hints students (CIFAR-10 trained teacher, TinyImageNet student). \textbf{(a)} Feature stability (average cosine similarity between clean images and their corrupted versions) measured over first 20,000 steps of training.  Feature stability of teacher shown as horizontal dashed line.  \textbf{(b)} PGR measured over each corruption type.}
  \label{fig:fs}
\end{figure}

\paragraph{Comparison with activation matching}
To isolate the effect of distilling representational structure, we compare BIRD to Hints, which supervises the student via linear mapping mean-squared error on the same representation layers. Despite identical teacher-student pairs and alignment points, BIRD achieves higher robustness in every setting and corruption category (Figure \ref{fig:fs}). This suggests that representation structure (captured via CKA) encodes more generalizable behavioral information than raw activation values. We conclude that BIRD’s success stems not just from where it supervises, but from how it supervises.

\begin{figure}[t!]
  \centering
  \includegraphics[width=0.9\linewidth]{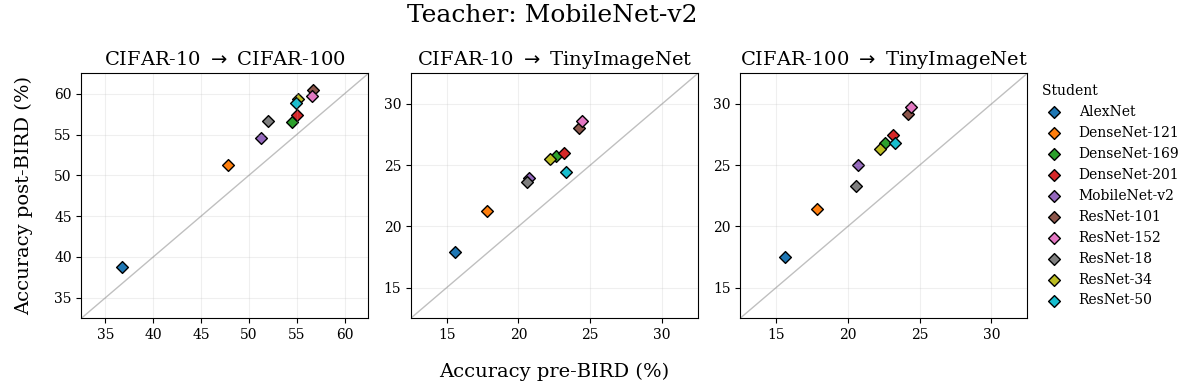}
  \caption{OOD robustness of student models before and after BIRD, guided by a MobileNetV2 teacher trained on a lower-complexity dataset. BIRD improves robustness across all student capacities, including models up to 25$\times$ larger than the teacher.}
  \label{fig:weak_teacher}
\end{figure}

\paragraph{Scaling to larger student models}
We next test whether BIRD generalizes to student models of higher capacity than the teacher. Fixing a MobileNetV2 teacher trained on CIFAR-10, we apply BIRD to a series of student architectures of varying capacity. As shown in Figure~\ref{fig:weak_teacher}, robustness improves across all model sizes, including a 22.4\% PGR for the ResNet-152 student, despite it being 25$\times$ larger than the teacher. These results confirm BIRD’s effectiveness for weak-to-strong behavior transfer, even in extreme capacity mismatches.
As is common in robust learning, improvements in OOD robustness are sometimes accompanied by minor reductions in clean accuracy. Detailed clean-vs-corruption breakdowns and per-seed results are provided in Appendix~\ref{app:robust_transfer}.

\section{What makes a good teacher for behavior transfer?}
\label{sec:explaining_transfer}

The preceding experiments demonstrated BIRD's ability to transfer robust behavior across datasets and architectures. 
Here, we investigate the conditions under which this transfer is most successful. 
Specifically, we want to understand what makes a teacher a good source for behavior transfer.

\paragraph{Setup and analysis framework}
Our analysis is based on the hypothesis that transfer success depends on two measurable properties of the teacher's representation space:
\begin{itemize}[topsep=0pt,leftmargin=15pt,parsep=0pt]
    \item \textbf{Task relevance:} the degree to which the teacher's representations are informative for the student’s downstream task.
    \item \textbf{Behavioral relevance:} the extent to which the teacher's representations support robust, aligned behavior.
\end{itemize}

To test this, we construct a pool of 144 teacher models varying along these two axes. Teachers are trained on one of three datasets ($\mathcal{D}_{teacher} \in \{$CIFAR-10, CIFAR-100, TinyImageNet$\}$) to manipulate task relevance. Robustness is varied by randomly selecting subsets of ImageNet-C corruptions to include during training, manipulating behavioral relevance. Teachers span four architectures: AlexNet \citep{alexnet}, ResNet18 \citep{he2016deep}, DenseNet121 \citep{huang2017densely}, and MobileNetV2 \citep{sandler2018mobilenetv2}.

Each teacher is used to supervise a ResNet50 student trained on one of three datasets ($\mathcal{D}_{student} \in \{$CIFAR-10, CIFAR-100, TinyImageNet$\}$), resulting in 432 teacher-student pairs. Robustness transfer is evaluated as in Section~\ref{sec:bird_robust_transfer}. To quantify the properties of each teacher’s representation space, we compute the following metrics:

\paragraph{Task relevance}
We train linear probes on the teacher's representation using clean data from $\mathcal{D}_{student}$ and measure:
\begin{itemize}[topsep=0pt,leftmargin=15pt,parsep=0pt]
    \item \textit{Probing accuracy:} classification accuracy of the linear probe~\citep{alain2016understanding} on held-out student data.
    \item \textit{Complementary knowledge:} fraction of student samples correctly classified by the teacher's probe but not by a probe trained on the student’s own representation~\citep{roth2023fantastic}.
\end{itemize}

\paragraph{Behavioral relevance}
We aggregate the $\gamma$-robust usefulness of each feature \citep{ilyas2019adversarial}, which measures whether features retain predictive value under corruptions. 
Full computation details are provided in Appendix~\ref{app:explaining}.

\paragraph{Explaining transfer success}
For each student dataset, we fit a linear model to predict accuracy on the corrupted test set using the three metrics described above.
We report $R^2$ values to quantify how much variance in behavior transfer can be explained by interpretable properties of the teacher's representation space.

\begin{figure}[t!]
  \centering
  \includegraphics[width=0.71\linewidth]{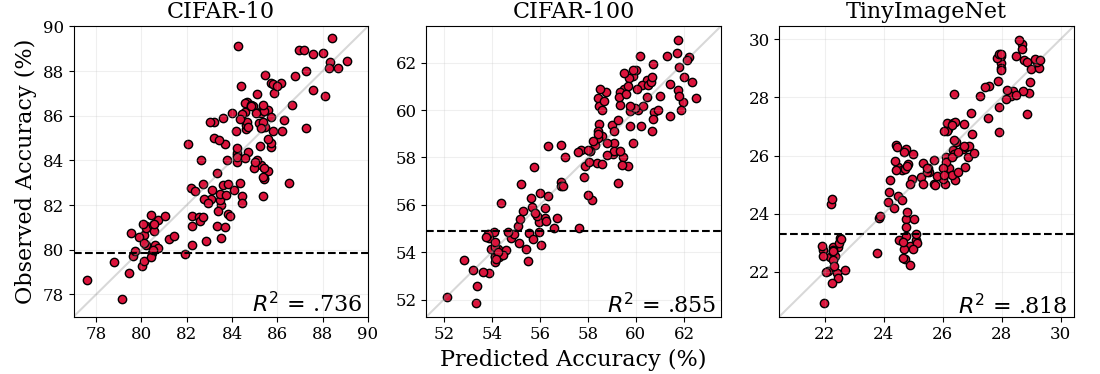}
  \caption{Variance in robust accuracy explained by linear models trained on teacher representation properties. Simple, interpretable properties of the teacher's representation space explain the majority of transfer outcomes.  Horizontal lines indicate student performance before BIRD.}
  \label{fig:r2}
\end{figure}

\paragraph{Results}
Impressively, these simple, interpretable properties explain the vast majority of transferred robustness.
Figure~\ref{fig:r2} shows the proportion of variance in robustness transfer explained by linear models trained on the three teacher metrics introduced in the previous section. 
We observe strong predictive power across all student datasets. 
For students trained on TinyImageNet, the model explains 81.8\% of the variance in PGR; for CIFAR-100, 85.5\%; and for CIFAR-10, 73.6\%. 
These results suggest that successful behavior transfer is not idiosyncratic, but can be anticipated using simple, interpretable properties of the teacher's representations.

Across all student datasets, the most predictive single factor is the behavioral relevance of the teacher's representation space, as measured by aggregated $\gamma$-robust usefulness. 
In each case, this factor alone explains more than 50\% of the variance in PGR. 
Task relevance metrics (probing accuracy and complementary knowledge) add additional explanatory power, especially when transferring to more complex student datasets like TinyImageNet.

\paragraph{Implications for teacher selection}
These findings provide actionable guidance for selecting or training effective teacher models. 
When choosing among candidate teachers, we recommend prioritizing those with high behavioral relevance, even if they were trained on a different dataset or have limited task overlap with the student. 
In low-resource settings, training a weak model on a small, curated dataset that elicits robust behavior may be a more efficient strategy than attempting to robustly train a large model directly.

\section{Generalizing BIRD beyond robustness}
\label{sec:lang}

To assess the generality of BIRD, we evaluate it in a distinct setting from vision-based robustness: weak-to-strong generalization in natural language processing (NLP) using soft-labels. This experiment examines whether BIRD can complement soft-label supervision and improve learning from small, aligned models in language tasks.

\paragraph{Setup and evaluation protocol}
We adopt the setup introduced by Burns et al., in which a small, aligned teacher model is fine-tuned on a target dataset and then used to supervise a larger student via soft predictions \citep{burns2023weak}. Specifically, we use GPT2-Small as the teacher and either GPT2-Medium or GPT2-Large as the student. All models are implemented via HuggingFace Transformers \citep{wolf2019huggingface}.

We consider three multiple-choice question answering datasets:
\begin{itemize}[topsep=0pt,leftmargin=15pt,parsep=0pt]
    \item \textbf{SciQ} \citep{SciQ}: science questions with factual answers.
    \item \textbf{BoolQ} \citep{clark2019boolq}: yes/no questions based on paragraph context.
    \item \textbf{Cosmos QA} \citep{huang-etal-2019-cosmos}: commonsense inference over short passages.
\end{itemize}

Each student is trained using one of two supervision strategies: (i) \emph{Soft-label distillation}: standard cross-entropy loss between student predictions and the teacher's soft probabilities, (ii) \emph{Soft-label + BIRD}: the same cross-entropy loss, augmented with BIRD's representation-structure loss computed over the final token embedding layer of teacher and student.
Following prior work, we compute accuracy on a held-out test set and report PGR relative to a ceiling set by direct fine-tuning of the student on ground-truth labeled target data.

\begin{table}[h]
\caption{Performance gap recovered (PGR) for GPT2-Medium and GPT2-Large models trained on soft-labels from GPT2-Small (Soft-Label) or soft-labels with BIRD (+BIRD).  Results on BoolQ are not shown: both methods provide no performance benefit over the weak teacher (0\% PGR). Reported results are averages over 3 seeds.}\label{tab:w_s}
\centering
\begin{tabular}{ccccc}
\toprule 
 Dataset & Student & \multicolumn{2}{c}{\% PGR ($\uparrow$)} \\
\cmidrule(lr){3-4}
 & & Soft-Label & +BIRD \\\midrule
\multirow{2}{*}{SciQ} & GPT2-Medium & 7.79 & \textbf{16.14} \\  
                       & GPT2-Large & 17.70 & \textbf{24.19}\\  
\midrule
\multirow{2}{*}{Cosmos QA} & GPT2-Medium & \textbf{47.00} & 42.76 \\  
                       & GPT2-Large & 65.51 & \textbf{68.02} \\  
\bottomrule \\
\end{tabular}
\end{table}

\paragraph{Results}
Table~\ref{tab:w_s} summarizes performance across all dataset and model combinations. In two of the six configurations, adding BIRD to soft-label distillation yields a noticeable improvement in behavior transfer.
For example, when supervising GPT2-Large on SciQ, BIRD increases PGR from 17.7\% to 24.2\%. On CosmosQA, BIRD improves PGR for GPT2-Large, though gains are more modest.

On BoolQ, neither soft-label distillation nor BIRD outperform the weak teacher. This suggests that not all aligned behaviors captured by small models generalize to more complex reasoning tasks or that representational distillation at the final layer may not be sufficient in this domain. Moreover, improvements from BIRD are more pronounced when the teacher and student differ significantly in size (e.g., GPT2-Small to GPT2-Large), consistent with its intended use in weak-to-strong transfer.

\paragraph{Implications for generalization}
These results show that BIRD can extend beyond vision and robustness settings, offering improvements in language tasks when combined with soft-label supervision. However, the effect is not universal. This highlights a broader opportunity: combining BIRD with additional alignment signals (e.g., multi-layer supervision or human preferences) may offer more consistent benefits in complex domains.

\section{Discussion}
\label{sec:discussion}

We introduce \acf{BIRD}, a simple and general framework for transferring aligned behavior by distilling the structure of a teacher model's internal representations. 
Unlike prior approaches, BIRD does not require shared training data, output space, or architecture between teacher and student. 
Our experiments in robust image classification show that BIRD consistently outperforms existing transfer methods, even when the teacher is significantly smaller and trained on a simpler dataset (Table~\ref{tab:robustness} and Figure~ \ref{fig:weak_teacher}).
We further identify three interpretable properties of the teacher's representation space (i.e., task relevance, behavioral relevance, and complementary knowledge) that strongly predict transfer success.
Finally, we demonstrate that BIRD extends beyond vision, offering complementary gains when applied alongside soft-label distillation in language models. 

\paragraph{Key takeaways}
BIRD contributes two key advances.
First, it establishes a flexible, drop-in mechanism for aligned behavior transfer that operates across domains, architectures, and label spaces by supervising over representation structure. 
This enables weak-to-strong generalization even when the teacher's training data is inaccessible or proprietary. 
Second, it offers a systematic analysis of what makes a good teacher for behavior transfer.
Our findings suggest that small, robust models can serve as effective alignment scaffolds for larger, unaligned models if their representations are behaviorally relevant.
BIRD is especially useful in settings where aligned behavior must be scaled without retraining large models or sharing sensitive data.

\paragraph{Limitations and future directions}
While promising, BIRD has some limitations. 
Our core analyses focus on robustness in vision, with only preliminary results in NLP. 
Extending BIRD to other behaviors (e.g., honesty) and modalities (e.g., multimodal reasoning) remains an open direction.
Second, transfer success is likely bounded by the student's capacity and task complexity; future work should quantify this saturation effect more precisely (Appendix \ref{app:saturation}). 
Third, we currently supervise alignment at a single layer selected by a heuristic. 
While our findings suggest that exact layer choice is not overly sensitive (Appendix~\ref{app:layer_selection}), future work may explore multi-layer or hierarchical extensions to capture deeper structural alignment.
Finally, our metrics for task and behavioral relevance are tailored to robust classification; in other domains, tools like linear tomography \citep{zou2023representation} or causal mediation analysis \citep{vig2020investigating} may be required to localize behavior-relevant representations.

\paragraph{Ethical considerations} 
BIRD is intended as a tool to improve human-AI alignment. However, its effectiveness depends on the teacher: if the teacher encodes biased, harmful, or misaligned behaviors, BIRD may transfer those behaviors to the student. Ensuring the integrity of teacher models is therefore essential for responsible deployment.



\newpage
{\small
\bibliographystyle{unsrt}
\bibliography{references}
}



\newpage

\appendix

\section{Code availability}

An implementation of BIRD will be made available as a public GitHub repository upon publication.

\section{Evaluating robust behavior transfer}
\label{app:robust_transfer}

\subsection{Out-of-distribution corruptions from the ImageNet-C benchmark}

\begin{figure}[!h]
  \centering
  \includegraphics[width=0.72\linewidth]{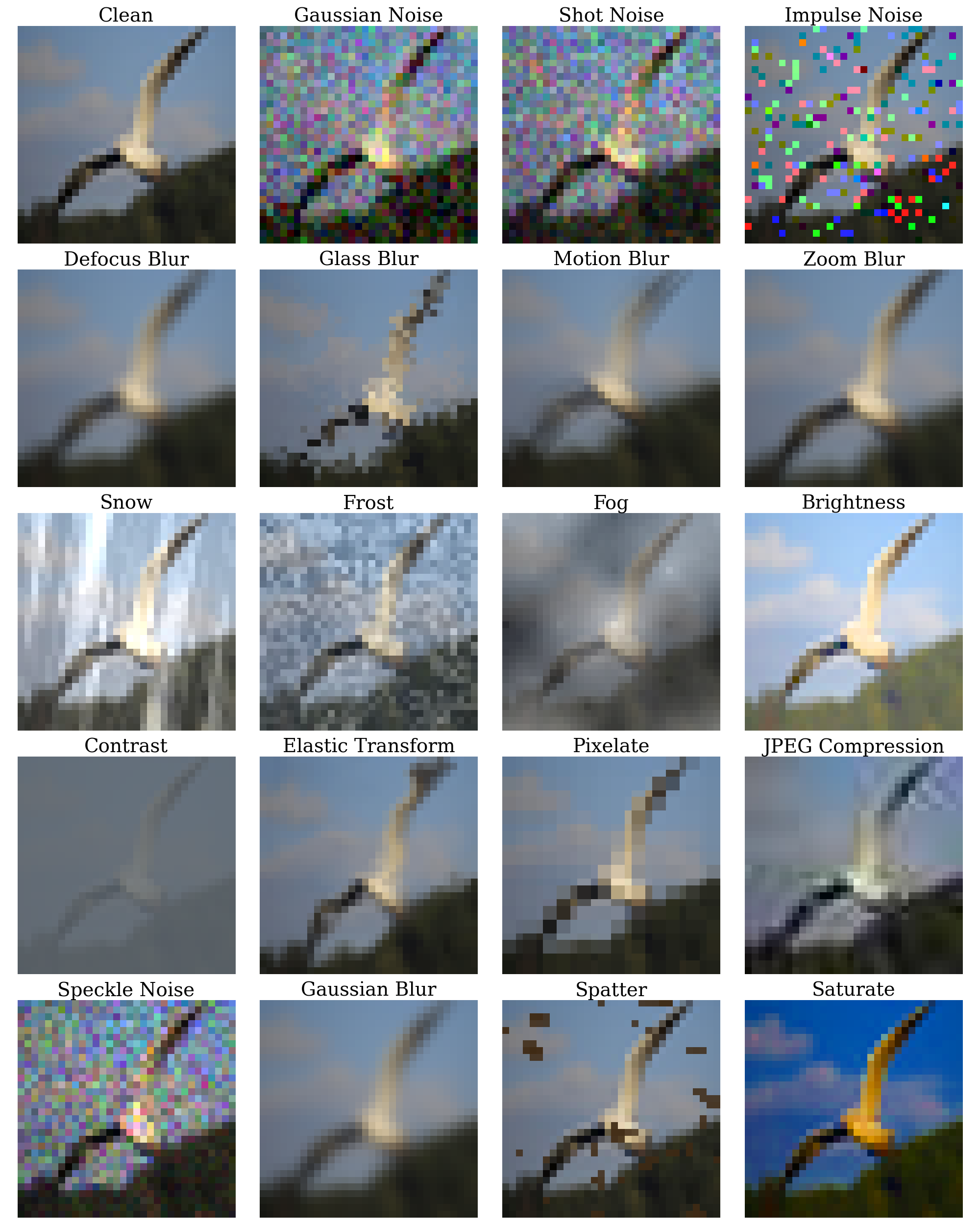}
  \caption{Sample CIFAR-10 image distorted with corruptions from the ImageNet-C benchmark \citep{hendrycks2019benchmarking}. ``Clean'' designates the original, uncorrupted image.  Corruptions are depicted at maximal severity.}
  \label{fig:app_corr_types}
\end{figure}

\begin{figure}[h!]
  \centering
  \includegraphics[width=\linewidth]{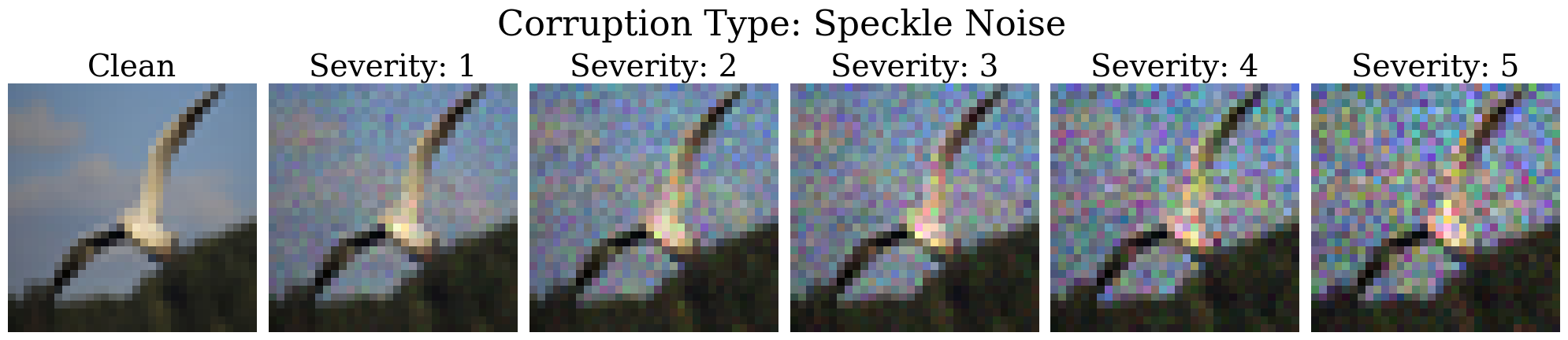}
  \caption{Sample CIFAR-10 image distorted with ``Speckle Noise'' corruption at severities 1-5.}
  \label{fig:app_corr_sevs}
\end{figure}

The ImageNet-C benchmark \citep{hendrycks2019benchmarking} is a standardized suite for evaluating the robustness of image classification models to common corruptions.
These image corruptions are algorithmically generated to simulate four different categories of real-world sources of image corruption (noise, blurring, weather effects, and digital effects).
These corruptions can be readily applied to images from datasets other than ImageNet (e.g., CIFAR-10, CIFAR-100, and TinyImageNet), as was done in this work.
There are 19 total corruption types, each from one of these four corruption categories (Figure \ref{fig:app_corr_types}).
Each corruption is additionally applied at five levels of severity, reflecting of the magnitude of the corruption (Figure \ref{fig:app_corr_sevs}).
In all experiments of Section \ref{sec:bird_robust_transfer}, teacher models are trained to be robust to 15 of these corruption types (``gaussian noise'', ``shot noise'', ``impulse noise'', ``defocus blur'', ``glass blur'', ``motion blur'', ``zoom blur'', ``snow'', ``frost'', ``fog'', ``brightness'', ``contrast'', ``brightness'', ``pixelate'', and ``jpeg compression'') and, after robust transfer, students are evaluated on the remaining 4 four held out corruption types (``speckle noise'', ``gaussian blur'', ``spatter'', and ``saturate''), one from each corruption category, to evaluated generalized robustness \citep{hendrycks2019benchmarking}.

\newpage
\subsection{Robust transfer baselines}

In Section \ref{sec:bird_robust_transfer}, we seek to compare robust behavior transfer with BIRD to robust transfer learning baselines that (1) solely rely on learning from clean data during the transfer learning task and (2) do not make inherent assumptions about the exact robust pre-training method.
These two criteria strongly reflect the behavior transfer paradigm that is studied in this paper: transferring behavior of an aligned model to a new, target domain using only the data that we already have available for the target domain.
To the best of our knowledge, the most relevant and highest performing methods that meet these criteria are robust transfer learning using continual learning strategies (LwF) \citep{shafahi2019adversarially}, Linear Probing (LP), Fine-tuning (FT), and Linear Probing followed by Fine-tuning (LP-FT) \citep{shafahi2019adversarially,kumar2022fine,nern2023transfer}.

\paragraph{Linear Probing (LP)} In LP, the robust, pre-trained feature extractor of a model is frozen while a new linear classification head is trained for the downstream task. This is a computationally efficient approach, often used when labeled data is scarce. Since the feature extractor is not updated, the robustness of the new model is largely dependent on the robustness of the features learned in pre-training \citep{shafahi2019adversarially,nern2023transfer}.

\paragraph{Fine-tuning (FT)} In FT, the entire model (i.e., the feature encoder and classification head) is updated as the model is trained on the new task or domain.  This provides more flexibility than LP, as the model is able to learn new features that support better clean performance on the new downstream task, but risks degrading the robustness of the original representations \citep{shafahi2019adversarially,nern2023transfer}.

\paragraph{Linear Probing followed by Fine-tuning (LP-FT)} LP-FT is performed in two steps.  First, a robust model's feature extractor is frozen and a new classification head for the target dataset is learned (LP).  Next, full-model fine-tuning is performed, updating both the model's feature extractor and new classification head.
Kumar et al. demonstrate that fine-tuning a model with a new, randomly initialized classification head can distort a model's learned features (ultimately contributing to worse OOD performance) and that these effects can be mitigated by first learning a classification head for the new task using LP \citep{kumar2022fine}.  Nern at al. employ this strategy in the context of transferring adversarial robustness in an effort to combine the benefits of LP (preserving robustness) and FT (improving performance) \citep{nern2023transfer}.

\paragraph{Learning without Forgetting (LwF)} Shahafi et al. suggest the use of LwF, a popular strategy used in lifelong learning tasks \citep{li2017learning}, as a method to preserve adversarial robustness when end-to-end fine-tuning a model for a new data distribution \citep{shafahi2019adversarially}. The approach augments the training objective with a loss that penalizes deviations between penultimate layer feature activations of the fine-tuned model and those of a robust source model. Unlike traditional LwF which distills from logits, this variant applies the loss directly to the penultimate feature layer, encouraging the student to retain robust internal representations even as it adapts to the target domain.

\subsection{Training details}

All experimental results for robust behavior transfer of Section \ref{sec:bird_robust_transfer} are reported over three seeds.
For each method that introduced additional hyperparameters, we instantiate that method with a range of hyperparameter values and report best performance achieved.
Specifically, for LwF \citep{shafahi2019adversarially} and Hints \citep{romero2014fitnets}, we tune the strength of the loss that penalizes divergence in feature activation values.
For FT, LP, FT-LP, and LwF, we additionally trained models with three unique initial learning rates and always report best performance.  We use a fixed optimization strategy in all methods (stochastic gradient descent with momentum and weight decay) and decay the learning rate with a cosine decay schedule over the course of training.

During robust transfer, all student models were exclusively trained on clean images.
Early stopping was performed using a clean validation set.

All models were trained using an Nvidia RTX 3090 GPU.
Teacher and student model training times were on the order of a few hours.

\subsection{Layer selection}
\label{app:layer_selection}

To select the distillation layers for the teacher and student models, we applied a simple heuristic based on representation measurements reported in Section \ref{sec:explaining_transfer}. Specifically, over multiple candidate layers associated with natural transition points in each architecture (e.g., the end of a ResNet, MobileNet, or DenseNet block), we computed two metrics (linear-probing accuracy and $\gamma$-robust usefulness from the CIFAR-10 $\rightarrow$ CIFAR-100 transfer task) as proxies of task relevance and behavioral relevance of the representations from that layer, and selected the layer with a highest composite score as the distillation anchor.  This composite score was computed as the mean of these two metrics, after 0-1 scaling each metric over the candidate layers from a given network.

\begin{figure}[h]
  \centering
  \includegraphics[width=0.7\linewidth]{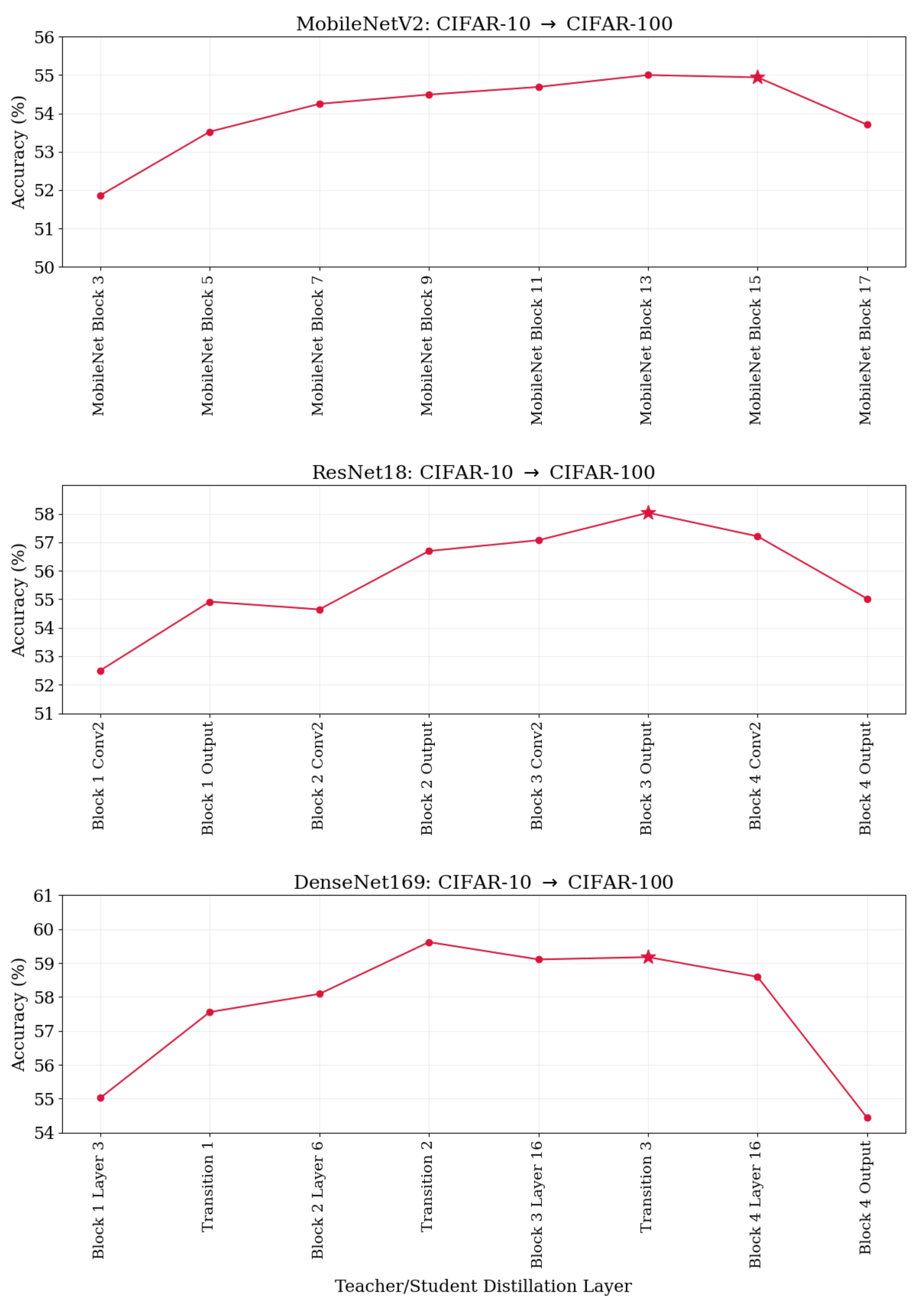}
  \caption{Robust accuracy of CIFAR-100 student after applying BIRD with differing distillation layers and a robust CIFAR-10 trained teacher.  Starred data points reflect the distillation layers used in the experiments of Section \ref{sec:bird_robust_transfer}.}
  \label{fig:layer_sel}
\end{figure}

While this procedure may not guarantee optimal performance for every BIRD-tuned model, it provided a computationally efficient proxy for selecting meaningful layers without exhaustively searching all possibilities. In practice, this strategy yielded strong results across architectures, and further improvements are likely achievable through layer-level validation when resources permit. For consistency and simplicity, we fixed the selected layer within each model family (e.g., always distilling from Block 3 outputs in ResNets) and transfer learning configuration (i.e., CIFAR-10 $\rightarrow$ CIFAR-100, CIFAR-10 $\rightarrow$ TinyImageNet, and CIFAR-100 $\rightarrow$ TinyImageNet). A broader evaluation of alternative layer choices is shown in Figure \ref{fig:layer_sel}.  Consistently, we observe that distilling representation structure at layers between the middle and end of the network offers best performance in this robust behavior transfer task.  Small deviations (e.g., a few hidden layers) away from the optimal distillation layer does not drastically reduce performance.

\subsection{Expanded results}

\begin{table}[t]
\caption{Per-seed accuracy (\%) of MobileNetV2 (MN2), ResNet18 (RN18), and DenseNet169 (DN169) models after behavior transfer using clean data from  CIFAR-10 (C10), CIFAR-100 (C100), and TinyImageNet (TIN).  Values reported are accuracy over all clean and corrupted images (for held-out corruption types ``speckle-noise'', ``gaussian-blur'', ``spatter'', and ``saturate'' and corruption severities $1$-$5$) from the target test set.} \label{tab:robustness_seeds}
\centering
\begin{tabular}{lcc c c c c c c c}
\toprule
 Model & Source & Target & Seed & \multicolumn{6}{c}{Accuracy of Behavior Transfer Method ($\uparrow$)} \\
 \cmidrule(lr){5-10}
 & Data & Data & &  LP & FT & LP-FT & Hints & LwF &  BIRD \\
\midrule
\multirow{9}{*}{MN2}         & C10  & C100 & 0 & 10.95 & 51.38 & 50.44 & 51.63 & 52.64 & {54.57}  \\
                             & C10  & C100 & 1 & 10.93 & 50.86 & 46.25 & 51.48 & 52.11 & {54.91}  \\
                             & C10  & C100 & 2 & 10.97 & 51.13 & 46.06 & 51.49 & 51.88 & {54.84}  \\
                             & C10  & TIN & 0 & 5.25  & 20.08 & 19.79 & 21.48 & 20.51 & {23.97}  \\
                             & C10  & TIN & 1 & 5.25  & 19.90 & 17.12 & 21.21 & 20.44 & {24.29}  \\
                             & C10  & TIN & 2 & 5.22  & 20.03 & 17.45 & 21.13 & 20.62 & {24.08}  \\
                             & C100 & TIN & 0 & 18.81 & 20.69 & 23.45 & 21.39 & 23.11 & {25.03}  \\
                             & C100 & TIN & 1 & 18.87 & 20.76 & 23.68 & 20.76 & 22.97 & {25.06}  \\
                             & C100 & TIN & 2 & 18.82 & 20.52 & 23.44 & 21.64 & 23.46 & {24.99}  \\
\midrule
\multirow{9}{*}{RN18}       & C10 & C100 & 0 & 16.94 & 51.69  & 52.39 & 52.39 & 55.34 & {57.37}  \\
                            & C10 & C100 & 1 & 16.92 & 52.08  & 49.46 & 52.03 & 54.60 & {57.57}  \\
                            & C10 & C100 & 2 & 16.94 & 52.09  & 50.02 & 52.18 & 56.33 & {57.23}  \\
                            & C10   & TIN & 0 & 7.26  & 20.01 & 20.07 & 20.82 & 22.11 & {23.12}  \\
                            & C10   & TIN & 1 & 7.24  & 19.87 & 19.01 & 20.96 & 21.82 & {23.82}  \\
                            & C10   & TIN & 2 & 7.26  & 20.41 & 19.14 & 20.97 & 22.59 & {23.86}  \\
                            & C100  & TIN & 0 & 20.97  & 20.98 & 23.88 & 20.73 & 24.22 & {24.27}  \\
                            & C100  & TIN & 1 & 20.89  & 20.59 & 23.51 & 20.60 & 25.10 & {24.53}  \\
                            & C100  & TIN & 2 & 20.98  & 20.68 & 23.58 & 20.80 & 24.13 & {24.67}  \\
\midrule
\multirow{3}{*}{DN169}       & C10 & C100 & 0 & 23.97 & 56.08 & 54.92 & 55.14 & 56.44 & {59.27}  \\
                             & C10 & C100 & 1 & 23.88 & 55.94 & 52.76 & 54.45 & 57.21 & {59.50}  \\
                             & C10 & C100 & 2 & 23.91 & 55.51 & 52.48 & 55.18 & 57.11 & {58.36}  \\
                             & C10 & TIN & 0 & 10.60 & 23.21 & 22.56 & 22.62 & 24.15 & {25.26}  \\
                             & C10 & TIN & 1 & 10.70 & 23.86 & 20.55 & 22.68 & 24.28 & {25.28}  \\
                             & C10 & TIN & 2 & 10.67 & 23.10 & 20.47 & 22.76 & 23.99 & {25.21}  \\
                             & C100 & TIN & 0 & 23.54 & 22.93 & 24.77 & 22.67 & 26.70 & {27.27}  \\
                             & C100 & TIN & 1 & 23.55 & 23.24 & 25.12 & 22.86 & 25.32 & {27.64}  \\
                             & C100 & TIN & 2 & 23.58 & 23.41 & 24.70 & 22.72 & 26.41 & {27.48}  \\
\bottomrule 
\end{tabular}
\end{table}

\begin{table}[t]
\caption{\textit{Clean} accuracy (\%) of MobileNetV2 (MN2), ResNet18 (RN18), and DenseNet169 (DN169) models after behavior transfer using clean data from  CIFAR-10 (C10), CIFAR-100 (C100), and TinyImageNet (TIN). Reported results are averaged over 3 seeds.} \label{tab:transfer_clean}
\centering
\begin{tabular}{lcc c c c c c c c}
\toprule
 Model & Source & Target & \multicolumn{7}{c}{Clean Accuracy of Behavior Transfer Method ($\uparrow$)} \\
 \cmidrule(lr){4-10}
 & Data & Data & None & LP & FT & LP-FT & Hints & LwF &  BIRD\\
\midrule
\multirow{3}{*}{MN2} & C10  & C100 & 74.27 & 12.28 & 75.34 & 70.67 & 74.00 & 75.14 & {71.89}  \\
                     & C10  & TIN  & 55.30 & 6.36  & 53.44 & 48.32 & 55.40 & 53.05 & {52.63}  \\
                     & C100 & TIN  & 55.30 & 24.75 & 53.97 & 48.52 & 55.11 & 52.96 & {53.29}  \\
\midrule
\multirow{3}{*}{RN18} & C10 & C100  & 75.96 & 19.07 & 76.15 & 72.61 & 75.60 & 75.99 & {73.92}  \\
                      & C10 & TIN   & 55.28 & 9.13  & 55.04 & 47.64 & 55.03 & 55.06 & {52.50}  \\
                      & C100 & TIN  & 55.28 & 28.18 & 55.50 & 50.36 & 54.91 & 53.73 & {52.33}  \\
\midrule
\multirow{3}{*}{DN169} & C10 & C100 & 78.21 & 27.74 & 79.75 & 76.16 & 78.21 & 77.95 & {75.48}  \\
                       & C10 & TIN  & 59.00 & 13.64 & 60.28 & 55.25 & 58.77 & 58.49 & {55.11}  \\
                       & C100 & TIN & 59.00 & 31.70 & 59.88 & 50.37 & 58.72 & 56.57 & {55.48}  \\
\bottomrule \\
\end{tabular}
\end{table}

\begin{figure}[h]
  \centering
  \includegraphics[width=0.8\linewidth]{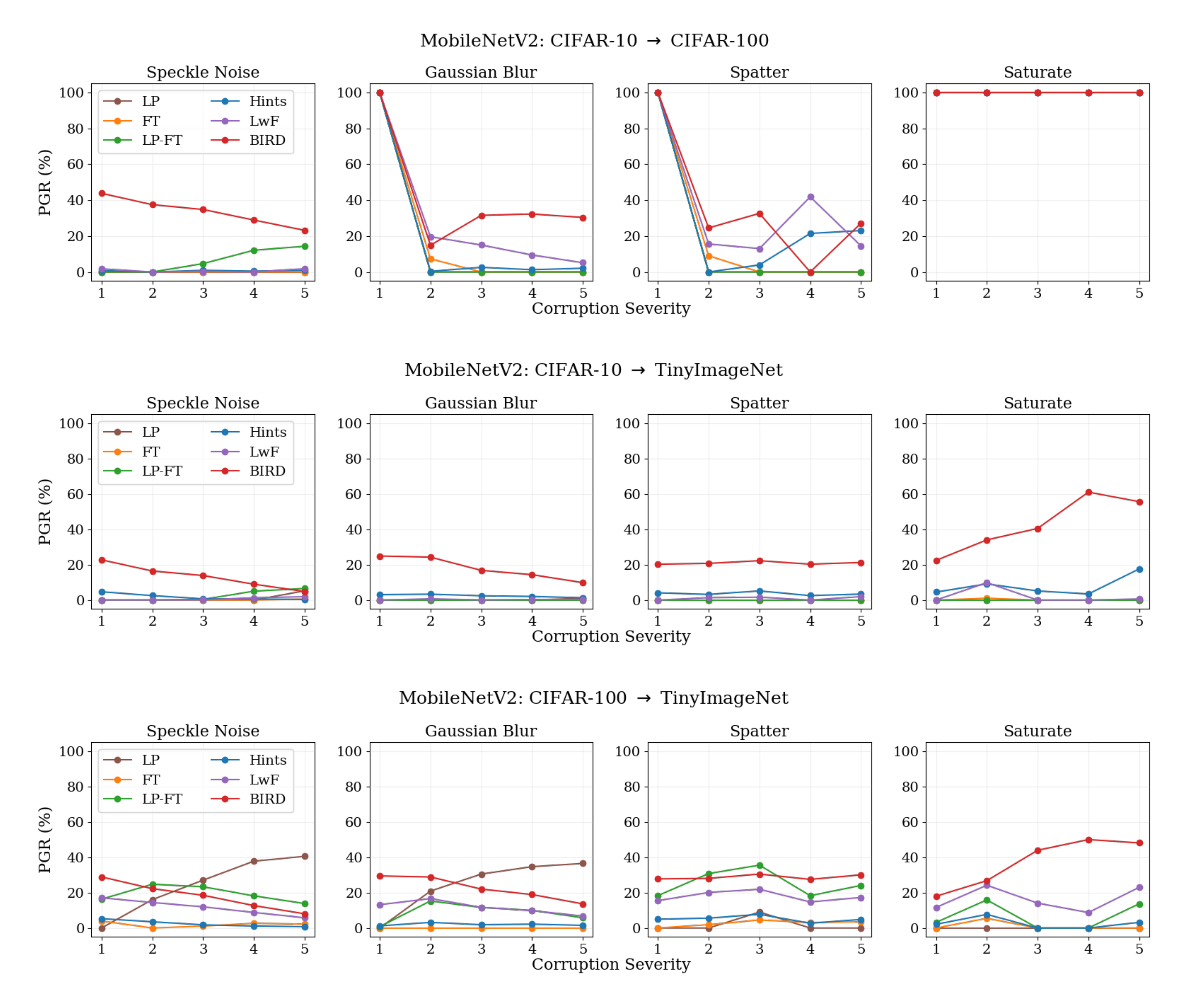}
  \caption{Performance gap recovered (PGR) of robust transfer methods applied to MobileNetV2, by test corruption type and severity.  PGR of 100\% indicates that the accuracy of the model at that corruption severity was greater than or equal to that of the ``Robust'' model trained on corruption-augmented data.}
  \label{fig:mn_pgr_at_sev}
\end{figure}

\begin{figure}[!t]
  \centering
  \includegraphics[width=0.8\linewidth]{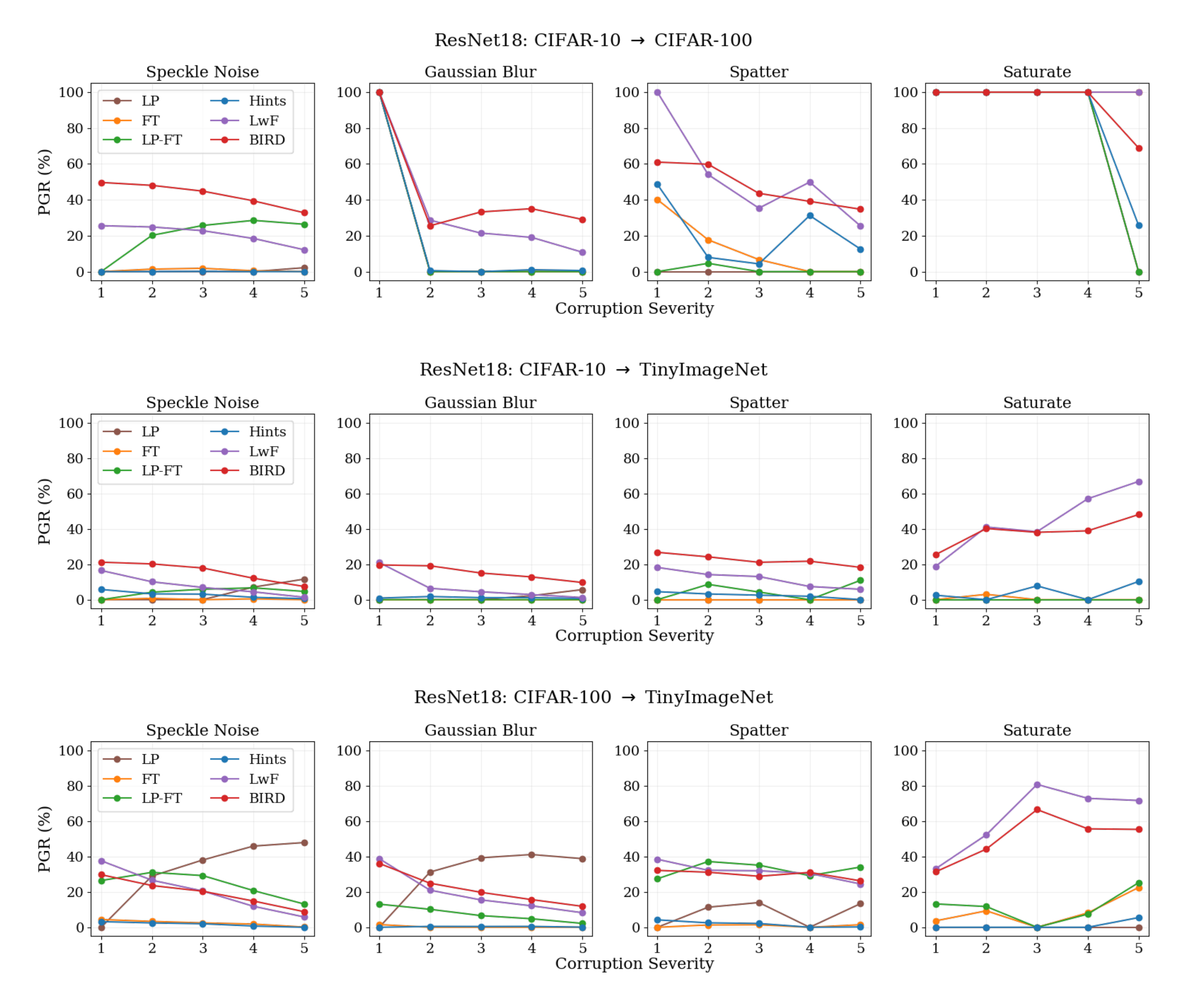}
  \caption{Performance gap recovered (PGR) of robust transfer methods applied to ResNet18, by test corruption type and severity.  PGR of 100\% indicates that the accuracy of the model at that corruption severity was greater than or equal to that of the ``Robust'' model trained on corruption-augmented data.}
  \label{fig:rn_pgr_at_sev}
\end{figure}

\begin{figure}[!b]
  \centering
  \includegraphics[width=0.8\linewidth]{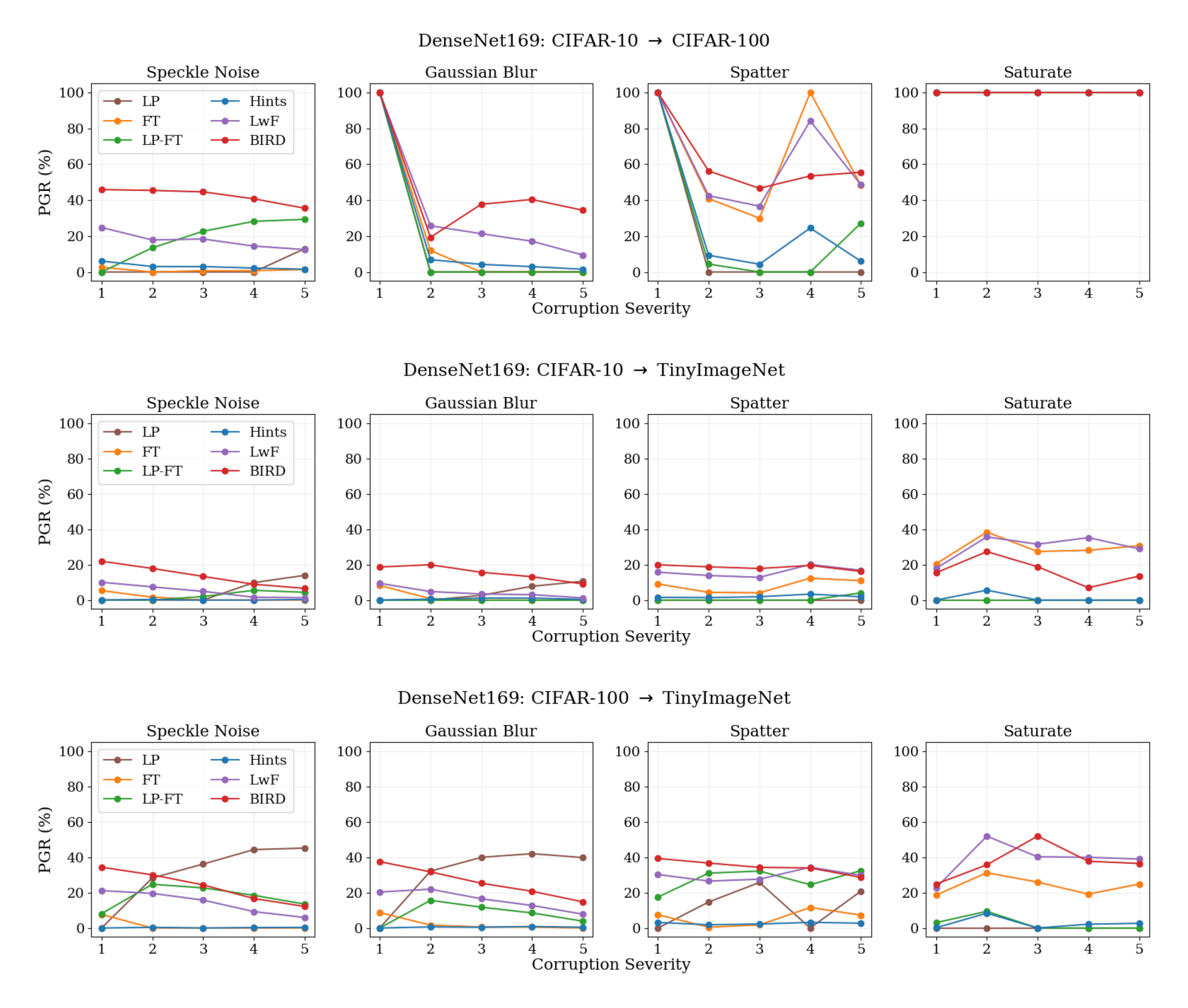}
  \caption{Performance gap recovered (PGR) of robust transfer methods applied to DenseNet169, by test corruption type and severity.  PGR of 100\% indicates that the accuracy of the model at that corruption severity was greater than or equal to that of the ``Robust'' model trained on corruption-augmented data.}
  \label{fig:dn_pgr_at_sev}
\end{figure}

Individual seed results for each robust transfer method evaluated in Section \ref{sec:bird_robust_transfer} are provided in Table \ref{tab:robustness_seeds}.
Clean accuracy of each model is reported in Table \ref{tab:transfer_clean}.
Figures \ref{fig:mn_pgr_at_sev}, \ref{fig:rn_pgr_at_sev}, and \ref{fig:dn_pgr_at_sev} show PGR of all methods for each corruption type and severity.

\subsection{Performance over distribution of classes}

\begin{figure}[h]
  \centering
  \includegraphics[width=\linewidth]{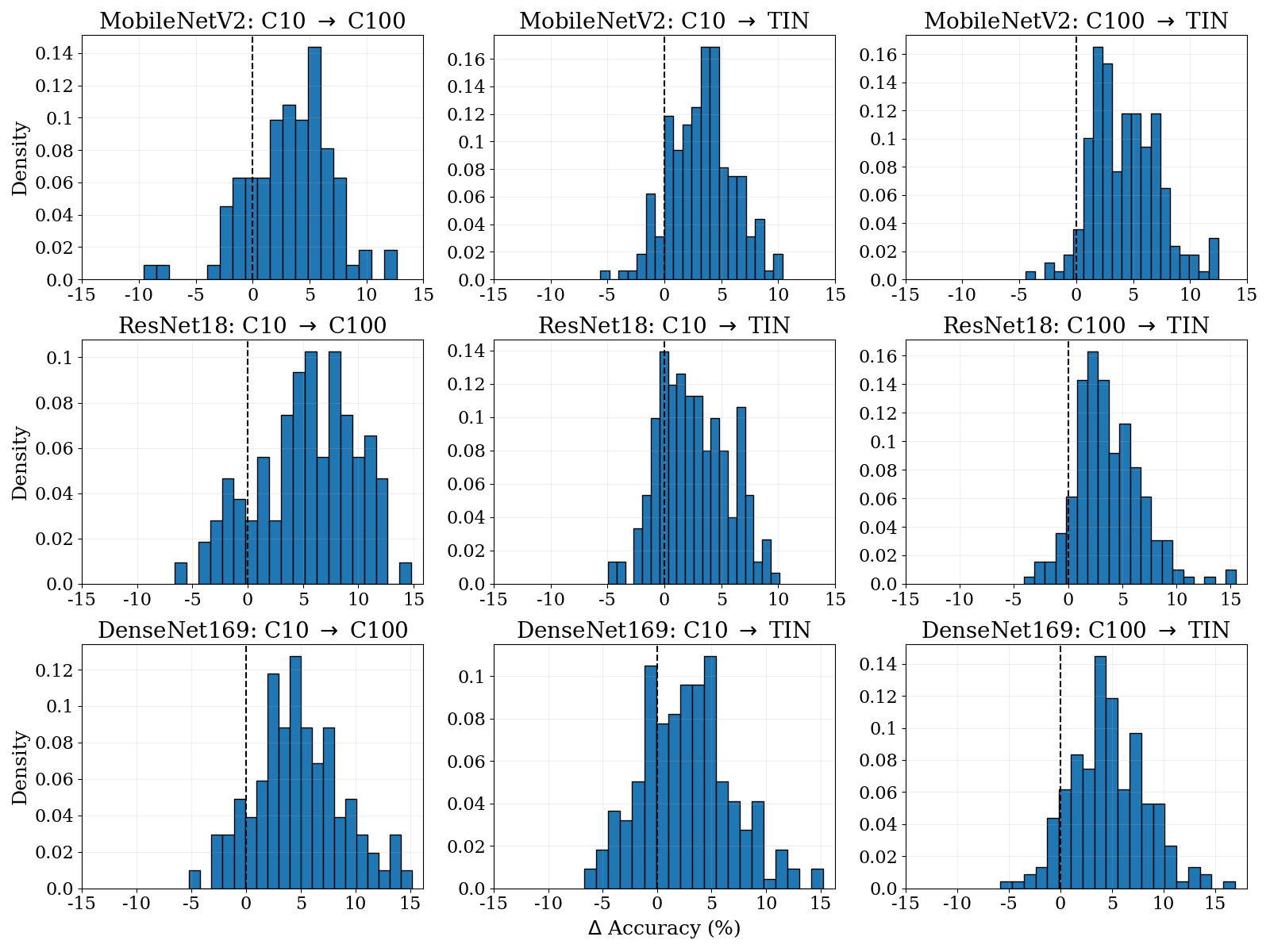}
  \caption{Change in per-class robust accuracy after fine-tuning with BIRD for each configuration of Section \ref{sec:bird_robust_transfer}.  Each bar reflects the proportion of classes that realized a given change in accuracy.}
  \label{fig:class_improvement_hists}
\end{figure}

Figure \ref{fig:class_improvement_hists} visualizes within-class accuracy change distributions after applying BIRD.
We observe that improvements in robustness are broadly distributed across classes, indicating that the gains are not limited to the few categories of the target dataset that are most similar to those of the teacher's source datset.

\section{Learning from weak teachers}
\label{app:learning_from_the_weak}

Quantitative results associated with Figure \ref{fig:weak_teacher} (performing behavior transfer from a MobileNetV2 teacher to student models of vaying capacity) are provided in Table \ref{tab:weak_teacher_numbers}. 

\begin{table}[t]
\caption{Accuracy over all test data (clean and corrupted) of 10 student models of varying capacity. \textbf{None:} student is trained from scratch on clean target training data. \textbf{BIRD:} Pre-trained, non-robust student is fine-tuned with a MobileNetV2 teacher using BIRD.  Accuracy on clean test data only is parenthesized.} \label{tab:weak_teacher_numbers}
\centering
\begin{tabular}{lcc c c}
\toprule
 Student Model & Source Data & Target Data & \multicolumn{2}{c}{Accuracy ($\uparrow$)} \\
 \cmidrule(lr){4-5}
 & & & None &  BIRD \\
\midrule
\multirow{3}{*}{MobileNetV2}    & C10  & C100 & 51.31 (74.27) & 54.57 (71.60) \\
                        & C10  & TIN  & 20.74 (55.30) & 23.97 (52.54) \\
                        & C100 & TIN  & 20.74 (55.30) & 25.03 (53.45) \\
\midrule
\multirow{3}{*}{AlexNet}    & C10  & C100 & 36.72 (56.69) & 38.79 (57.56) \\
                            & C10  & TIN  & 15.59 (32.78) & 17.92 (33.59) \\
                            & C100 & TIN  & 15.59 (32.78) & 17.51 (33.13) \\
\midrule
\multirow{3}{*}{ResNet18}    & C10  & C100 & 52.03 (75.96) & 56.61 (73.47) \\
                             & C10  & TIN  & 20.56 (55.28) & 23.65 (52.26) \\
                             & C100 & TIN  & 20.56 (55.28) & 23.32 (52.70) \\
\midrule
\multirow{3}{*}{ResNet34}    & C10  & C100 & 55.09 (76.30) & 59.37 (74.23) \\
                             & C10  & TIN  & 22.21 (56.78) & 25.48 (53.66) \\
                             & C100 & TIN  & 22.21 (56.78) & 26.30 (54.00) \\
\midrule
\multirow{3}{*}{ResNet50}    & C10  & C100 & 54.89 (77.24) & 58.84 (75.66) \\
                             & C10  & TIN  & 23.29 (59.84) & 24.46 (56.22) \\
                             & C100 & TIN  & 23.29 (59.84) & 26.80 (56.73) \\
\midrule
\multirow{3}{*}{ResNet101}    & C10  & C100 & 56.67 (76.47) & 60.43 (74.75) \\
                              & C10  & TIN  & 24.22 (59.81) & 28.07 (56.27) \\
                              & C100 & TIN  & 24.22 (59.81) & 29.20 (56.76) \\
\midrule
\multirow{3}{*}{ResNet152}    & C10  & C100 & 56.55 (74.80) & 59.75 (73.64) \\
                              & C10  & TIN  & 24.41 (58.99) & 28.63 (56.43) \\
                              & C100 & TIN  & 24.41 (58.99) & 29.77 (56.50) \\
\midrule
\multirow{3}{*}{DenseNet121}    & C10  & C100 & 47.79 (73.42) & 51.27 (70.36) \\
                                & C10  & TIN  & 17.84 (50.55) & 21.22 (52.62) \\
                                & C100 & TIN  & 17.84 (50.55) & 21.44 (52.31) \\
\midrule
\multirow{3}{*}{DenseNet169}    & C10  & C100 & 54.51 (78.21) & 56.52 (73.36) \\
                                & C10  & TIN  & 22.59 (59.00) & 25.74 (54.03) \\
                                & C100 & TIN  & 22.59 (59.00) & 26.79 (54.58) \\
\midrule
\multirow{3}{*}{DenseNet201}    & C10  & C100 & 55.07 (79.23) & 57.37 (74.38) \\
                                & C10  & TIN  & 23.16 (59.11) & 25.97 (54.66) \\
                                & C100 & TIN  & 23.16 (59.11) & 27.48 (55.60) \\
\bottomrule \\
\end{tabular}
\end{table}

\section{Explaining robust behavior transfer}
\label{app:explaining}

\subsection{Quantifying behavioral relevance of a teacher's feature space}

To quantify the behavioral relevance of a teacher model’s representations, we adapt the notion of $\gamma$-robust useful features from Ilyas et al. \citep{ilyas2019adversarial}, which defines a feature as robustly useful if it remains predictive of the true label even under allowable input perturbations.
While the original formulation applies to binary classification under adversarial perturbations, we extend this idea to the multi-class setting and apply it to common image corruptions (rather than adversarial attacks).

We compute a behavioral relevance score for a given model layer using the following procedure:
\begin{enumerate}[topsep=0pt,leftmargin=15pt,parsep=0pt]
    \item Compute class-wise $\gamma$-robust usefulness: For each feature dimension in the representation space, and for each class $c$ in the target dataset, we binarize the class labels such that samples from class $c$ are labeled as $+1$ and all other samples as $-1$. We then evaluate the feature’s correlation with this binary labeling across target dataset images distorted with corruptions from the ImageNet-C benchmark. This gives a robustness-aware measure of the feature’s predictive power for each class.
    \item Feature-level aggregation: For each feature dimension, we compute the $90^{th}$ percentile of its $\gamma$-robust usefulness scores across all classes. This choice reflects the intuition that a useful feature may be strongly predictive of a subset of classes.
    \item Layer-level aggregation: Finally, we aggregate over all feature dimensions by taking the median of the feature-level scores. This yields a single scalar score representing the typical $\gamma$-robust usefulness of features in the layer of interest.
\end{enumerate}

This metric allows us to compare candidate teacher layers by how robustly informative their features are, even in the presence of corruptions, and serves as a proxy for their suitability in behavior transfer via BIRD.

\section{Robust transfer from under-expressive students}
\label{app:saturation}

\begin{figure}[h]
  \centering
  \includegraphics[width=\linewidth]{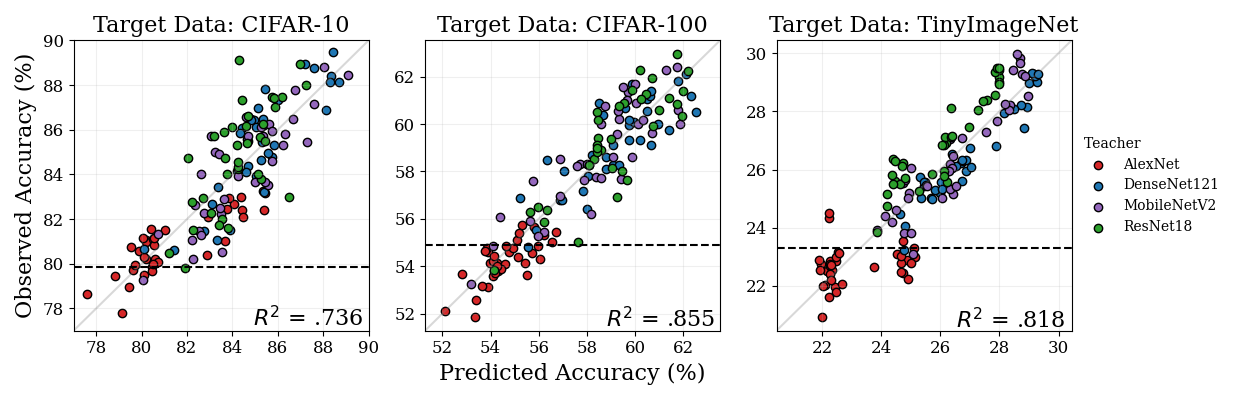}
  \caption{Variance in robust accuracy explained by linear models trained on teacher representation properties (as plotted in Figure \ref{fig:r2}), colored by teacher model architecture.}
  \label{fig:alexnet_failure_case}
\end{figure}

While our results demonstrate that robust behavior can be transferred from low-capacity teachers to higher-capacity students, there are signs of a saturation effect: when the gap between teacher and student capacity becomes too large, the effectiveness of behavior transfer diminishes.
In Section \ref{sec:explaining_transfer}, we seek to transfer robust behavior from AlexNet, ResNet18, DenseNet121, and MobileNetV2 teachers to ResNet50 students.
AlexNet teachers, which substantially underperform all other teachers in both clean and robust image classification, frequently fail to serve as useful teachers for transferring robust behavior, especially in the case if higher-complexity target datasets (Figure \ref{fig:alexnet_failure_case}).
This leads us to suggest that extremely weak or under-expressive teachers may lack sufficiently structured representations to guide meaningful behavior transfer via BIRD.

\clearpage
\newpage
\section{Extending soft-label, weak-to-strong generalization with BIRD}
\label{app:llm_training}

In Section \ref{sec:lang}, we study whether BIRD can complement soft-label based weak-to-strong generalization in language models on three simple datasets: SciQ \citep{SciQ}, BoolQ \citep{clark2019boolq}, and Cosmost QA \citep{huang-etal-2019-cosmos}.
Our implementation builds on OpenAI's publicly available github repository (\url{https://github.com/openai/weak-to-strong}).
We use identical hyperparameters as set by default for each training configuration in this repository, as we found that these provided consistently good results for baseline soft-label learning.
For BIRD, representation loss weight ($\alpha$ in Equation \ref{eq:bird_loss}), was selected based on performance from five psuedo-randomly selected values.  GPT2-Small and GPT2-Medium models were trained using a Nvidia RTX 3090 GPU.  GPT2-Large models were trained using a Nvidia A100 GPU. 

Results reported in Table \ref{tab:w_s} are averaged over three seeds, each operating on different train-test splits of the dataset.  Per-seed results are provided in Table \ref{tab:w_s_seeds}.

\begin{table}[t]
\caption{Per-seed performance gap recovered (PGR) results for GPT2-Medium and GPT2-Large models trained on soft-labels from GPT2-Small (Soft-Label) or soft-labels with BIRD (+BIRD).}
\label{tab:w_s_seeds}
\centering
\begin{tabular}{cccccc}
\toprule 
 Dataset & Student & Seed & \multicolumn{2}{c}{\% PGR ($\uparrow$)} \\
\cmidrule(lr){4-5}
 & & & Soft-Label & +BIRD \\\midrule
\multirow{6}{*}{SciQ} & GPT2-Medium & 0 & 33.33 & 0 \\  
                      & GPT2-Medium & 1 & 0 & 27.78 \\  
                      & GPT2-Medium & 2 & 12.24 & 26.53 \\  
                       & GPT2-Large & 0 & 27.06 & 38.82\\  
                       & GPT2-Large & 1 & 19.19 & 30.93\\  
                       & GPT2-Large & 2 & 9.86 & 2.82\\  
\midrule
\multirow{6}{*}{Cosmos QA} & GPT2-Medium & 0 & 16.94 & 21.29 \\
                           & GPT2-Medium & 1 & 95.15 & 100.00 \\
                           & GPT2-Medium & 2 & 28.89 & 7.00 \\
                           & GPT2-Large & 0 & 46.52 & 51.91 \\
                           & GPT2-Large & 1 & 100.00 & 100.00 \\
                           & GPT2-Large & 2 & 50.00 & 52.14 \\
\bottomrule \\
\end{tabular}
\end{table}


\end{document}